\documentclass{article}
\usepackage{ijcai17}

\usepackage{algorithm}
\usepackage{algpseudocode}
\usepackage{graphicx}
\usepackage{url} 
\usepackage{color}
\usepackage{subfigure}
\usepackage{amsmath}
\usepackage{comment}
\usepackage{amssymb}
\usepackage{times}
\usepackage[normalem]{ulem} 

\DeclareMathOperator*{\argmin}{arg\,min}
\newtheorem{defn}{Definition}[section]
\newtheorem{thm}{Theorem}[section]
\newtheorem{cor}{Corollary}[section]

\newcommand{\bi}{\begin{itemize}}
\newcommand{\ei}{\end{itemize}}
\newcommand{\BE}{\begin{enumerate}}
\newcommand{\EE}{\end{enumerate}}

\newcommand{\etal}{\mbox{\it et al.}}

\newtheorem{theorem}{Theorem}
\newtheorem{lemma}[theorem]{Lemma}

\title{Coarse-to-Fine Lifted MAP Inference in Computer Vision}

\author{Haroun Habeeb \and Ankit Anand \and Mausam \and Parag Singla\\ 
Indian Institute of Technology Delhi  \\
haroun7@gmail.com \and \{ankit.anand,mausam,parags\}@cse.iitd.ac.in
}
\begin{document}

\maketitle

\begin{abstract}
There is a vast body of theoretical research on {\em lifted inference} in probabilistic graphical models (PGMs). However, few demonstrations exist where lifting is applied in conjunction with top of the line applied algorithms. We pursue the applicability of lifted inference for computer vision (CV), with the insight that a globally optimal (MAP) labeling will likely have the same label for two symmetric pixels. The success of our approach lies in efficiently handling a distinct unary potential on every node (pixel), typical of CV applications. This allows us to lift the large class of algorithms that model a CV problem via PGM inference. We propose a generic template for coarse-to-fine (C2F) inference in CV, which progressively refines an initial coarsely lifted PGM for varying quality-time trade-offs. We demonstrate the performance of C2F inference by developing lifted versions of two near state-of-the-art CV algorithms for stereo vision and interactive image segmentation. We find that, against flat algorithms, the lifted versions have a much superior anytime performance, without any loss in final solution quality. 
\end{abstract}
\vspace{-1em}

\section{Introduction}
{\em Lifted inference} in probabilistic graphical models (PGMs) refers to the set of the techniques that carry out inference over groups of random variables (or states) that behave similarly \cite{jha&al10,kimmig&al15}. A vast body of theoretical work  develops a variety of lifted inference techniques, both exact (e.g., \cite{poole03,braz&al05,singla&domingos08,kersting12}) and approximate (e.g., \cite{singla&al14,broeck&niepert15}). Most of these works develop technical ideas applicable to generic subclasses of PGMs, and the accompanying experiments are aimed at providing first proofs of concepts. 
However, little work exists on transferring these ideas to the top {\em domain-specific} algorithms for real-world applications.

Algorithms for NLP, computational biology, and computer vision (CV) problems make heavy use of PGM machinery (e.g., \cite{blei&al03,friedman04,szeliski&al06}). But, they also include significant problem-specific insights to get high performance. Barring a handful of exceptions \cite{jernite&al15,nath&domingos16}, lifted inference hasn't been applied directly to such algorithms.

We study the potential value of lifting to CV problems such as image denoising, stereo vision, and image segmentation. Most CV problems are structured output prediction tasks, typically assigning a label to each pixel. A large class of solutions are PGM-based: they define a Markov Random Field (MRF) that has each pixel as a node, with unary potential that depends on pixel value, and pairwise neighborhood potentials that favor similar labels to neighboring pixels. 

We see three main challenges in applying existing lifted inference literature to these problems. First, most existing algorithms focus on computing marginals
\cite{singla&domingos08,kersting&al09,gogate&domingos11,niepert12,anand&al16,anand&al17} instead of MAP inference.  Second, among the algorithms performing lifted MAP~\cite{noessner&al13,mladenov&al14,sarkhel&al14,mittal&al14}, many of the algorithms focus on exact lifting. This breaks the kind of symmetries we need to compute since different pixels may not have exact same neighborhood. Third, the few algorithms that perform approximate lifting for MAP, e.g. \cite{sarkhel&al15}, can't handle a distinct unary potential on every node. This is essential for our application since image pixels take ordinal values in three channels.

In response, we develop an approximate lifted MAP inference algorithm which can effectively handle unary potentials. 
We initialize our algorithm by merging together pixels having the same order of top-$k$ labels based on the unary potential values. We then adapt an existing symmetry finding algorithm~\cite{kersting&al09} to discover groupings which also have similar neighborhoods. We refer to our groupings as {\em lifted pixels}. We impose the constraint that all pixels in a lifted pixel must be assigned the same label. Our approximate lifting reduces the model size drastically leading to significant time savings. Unfortunately, such approximate lifting could adversely impact solution quality. However, we vary the degree of approximation in symmetry finding to output a {\em sequence} of coarse-to-fine models with varying quality-time trade-offs. By switching between such models, we develop a coarse-to-fine (C2F) inference procedure applicable to many CV problems.

We formalize these ideas in a novel template for using lifted inference in CV. 
We test C2F lifted inference on two problems: stereo matching and image segmentation. We start with one of the best MRF-based solvers each for both problems -- neither of these are vanilla MRF solvers. Mozerov \& Weijer \shortcite{TSGO} use a {\em two-way} energy minimization to effectively handle occluded regions in stereo matching. Co-operative cuts \cite{kohli2013principled} for image segmentation use concave functions over a predefined set of pixel pairs to correctly segment images with sharp edges. We implement C2F inference on top of both these algorithms and find that C2F versions have a strong anytime behavior -- given any amount of inference time, they output a much higher quality (and are never worse) than their unlifted counterparts, and don't suffer any loss in the final quality. Overall, our contributions are:

\begin{enumerate}

\item We present an approximate lifted MAP algorithm that can efficiently handle a large number of distinct unary potentials.

\item We develop a novel template for applying lifted inference in structured prediction 
tasks in CV. We provide methods that output progressively finer approx. symmetries, leading to a C2F lifted inference procedure.

\item We implement C2F inference over a near state-of-the-art stereo matching algorithm, and one of the best  MRF-based image segmentation algorithms. We release our implementation for wider use by the community.\footnote{\em https://github.com/dair-iitd/c2fi4cv/}

\item We find that C2F has a much superior anytime behavior. For stereo matching it achieves 60\% better quality on average in time-constrained settings. For image segmentation C2F reaches convergence in 33\% less time.

\end{enumerate}
\vspace{-1em}
\section{Background}\label{sec:background}
\subsection{Computer Vision Problems as MRFs}
Most computer vision problems are structured output prediction problems and their PGM-based solutions often follow similar formulations. They cast the tasks into the problem of finding the lowest energy assignment over grid-structured MRFs (denoted by $G=(\mathcal{X},\gamma)$). The random variables in these MRFs are the set of pixels ${\cal X}$ in the input image. Given a set of labels $L:\{1,2,\ldots,|L|\}$, the task of structured output prediction is to label each pixel $X$ with a label from $L$. The MRFs have two kinds of potentials ($\gamma$) -- unary and higher-order. Unary potentials  are defined over each individual pixel, and usually incorporate pixel intensity, color, and other pixel features. Higher order potentials  operate over cliques (pairs or more) of neighboring pixels and typically express some form of {\em spatial homophily} -- ``neighboring pixels are more likely to have similar labels." While the general PGM structure of various tasks are similar, the specific potential tables and label spaces are task-dependent.

The goal is to find the MAP assignment over this MRF, which is equivalent to energy minimization (by defining energy as negative log of potentials). We denote the negative log of unary potentials by $\phi$, and that of higher-order potentials by $\psi$.\footnote{In the interest of readability, we say `potential' to mean `negative log of potential' in the rest of the paper.} Thus, energy of a complete assignment $\mathrm{\bf{x}}\in L^{|\mathcal{X}|}$ can be defined as:

\begin{eqnarray}
E(\mathrm{\bf{x}}) = \sum_{i \in 1..|\mathcal{X}|} \phi(x_i) + \sum_{j}\psi_j(\hat{x}_j)\quad 
\end{eqnarray}

Here $\hat{x}_j$ denotes the assignment $\mathbf{x}$ restricted to the set of variables in the potential $\psi_j$. And the output of the algorithm is the assignment $\mathrm{\mathbf{x}_{MAP}}$:
\begin{eqnarray}
\mathrm{\mathbf{x}_{MAP}}=\argmin\limits_{\mathbf{\mathrm{\bf{x}}}\in L^{|\mathcal{X}|}} E (\mathrm{\bf{x}})
\end{eqnarray}

The problem is in general intractable. Efficient approximations exploit special characteristics of potentials like submodularity \cite{jegelka11}, or use variants of graph cut or loopy belief propagation \cite{boykov2001fast,freeman&al00}. 

\subsection{Symmetries in Graphical Models}
Lifting an algorithm often requires computing a set of symmetries that can be exploited by that algorithm. For PGMs, two popular methods for symmetry computation are color passing for computing symmetries of variables \cite{kersting&al09}, and graph isomorphism for symmetries of states \cite{niepert12,bui&al13}. Since our work is based on color passing, we explain it in more detail.

Color passing for an MRF operates over a colored bipartite graph containing nodes for all variables and potentials, and each node is assigned a color. 
The graph is initialized as follows: all variables nodes get a common color; all potential nodes with exactly same potential tables are assigned a unique color. Now, in an iterative color passing scheme, in each iteration, each variable node passes its color to all neighboring potential nodes. The potential nodes store incoming color signatures in a vector, append their own color to it, and send the vector back to variable nodes. The variable nodes stack these incoming vectors. New colors are assigned to each node based on the set of incoming messages such that two nodes with same messages are assigned the same unique color. This process is repeated until convergence, i.e., no further change in colors. 

A coloring of the bipartite graph defines a partition of variable nodes such that all nodes of the same color form a partition element. Each iteration of color passing creates successively finer partitions, since two variable nodes, once assigned different colors, can never get the same color. 

\section{Lifted Computer Vision Framework} \label{sec:framework}

In this section, we will describe our generic template which can be used to lift a large class of vision applications including those in stereo, segmentation etc. Our template can be seen as transforming the original problem space to a reduced problem space over which the original inference algorithm can now be applied much more efficiently. Specifically, our description in this section is entirely \emph{algorithm independent}.

We will focus on MAP inference which is the inference task of choice for most vision applications (refer Section~\ref{sec:background}). 

The key insight in our formulation is based on the realization that pixels which are involved in the same (or similar) kinds of unary and higher order potentials, and have the same (or similar) neighborhoods, are likely to have the same MAP value. Therefore, if somehow we could discover such sets of pixels a priori, we could explicitly enforce these pixels to have the same value while searching for the solution, substantially reducing the problem size and still preserving the optimal MAP assignment(s). Since in general doing this exactly may lead to a degenerate network, we do it approximately. Hence trading-off speed for marginal loss in solution quality. The loss in solution quality is offset by resorting to coarse-to-fine inference where we start with a crude approximation, and gradually make it finer, to guarantee optimality at the end while still obtaining significant gains.
We next describe the details of our approach.

\subsection{Obtaining a Reduced Problem}
Consider an energy minimization problem over a PGM $G=(\mathcal{X},\gamma)$. Let $L=\{1,2,\cdots,|L|\}$ denote the set of labels over which variables in the set $\mathcal{X}$ can vary. Let $\mathcal{Y}^P=\{Y^P_1,Y^P_2,\cdots,Y^P_r\}$ denote a partition of $\mathcal{X}$ into $r$ disjoint subsets, i.e., $\forall k$, $Y_k^P \subseteq \mathcal{X}$, $Y^P_{k_1} \cap Y^P_{k_2} = \emptyset$ when $k_1 \neq k_2$, and $\cup_k Y^P_k = \mathcal{X}$. We refer to each $Y^P_k$ as a partition element. Correspondingly, let us define $\mathcal{Y}=\{Y_1,Y_2,\cdots,Y_r\}$ as a set of partition variables, where there is a one to one correspondence between partition elements and the partition variables and each partition variable $Y_k$ takes values in the set $L$. Let $part(X_i)$ denote the partition element to which $X_i$ belongs. Let $\hat{X}_j \subseteq \mathcal{X}$ denote a subset of variables. We say that a partition element $Y^P_k$ is {\emph represented} in the set $\hat{X}_j$ if $\exists X_i \in \hat{X}_j$ s.t. $part(X_j)=Y_k$. 

\noindent Given a subset of variables $\hat{X}_j$, let $\gamma_j(\hat{X}_j)$ be a potential defined over $\hat{X}_j$. Let $\hat{x}_j$ denote an assignment to variables in the set $\hat{X}_j$. Let $\hat{x}_j.elem(i)$ denote the value taken by a variable $X_i$ in $\hat{X}_j$. We say that an assignment $\hat{X}_j=\hat{x}_j$ {\emph respects} a partition $\mathcal{Y}^P$ if the variables in $\hat{X_j}$ belonging to the same partition element have the same label in $\hat{x}_j$, i.e., $part(X_i)=part(X_{i'}) \Rightarrow \hat{x}_j.elem(i)=\hat{x}_{j}.elem(i'), \forall X_i, X_{i'} \in \hat{X}_j$.
Next, we introduce the notion of a reduced potential.

\begin{defn}
Let $\mathcal{X}$ be a set of variables and let $\mathcal{Y}^P$ denote its partition. Given the potential $\gamma_j(\hat{X}_j)$, the reduced potential $\Gamma_{j}$ is defined to be the restriction of $\gamma_j(\hat{X}_j)$ to those labeling assignments of $\hat{X}_j$ which respect the partition $\mathcal{Y}^P$. Equivalently, we can define the reduced potential $\Gamma_{j}(\hat{Y}_j)$ over the set of partition variables $\hat{Y}_j$ which are represented in the set $\hat{X}_j$.
\end{defn}

For example, consider a potential $\gamma(X_1,X_2,X_3)$ defined over three Boolean variables. The table for $\gamma$ would have $8$ entries. Consider the partition $\mathcal{Y}^P=\{Y^P_1,Y^P_2\}$ where $Y^P_1=\{X_1,X_2\}$ and $Y^P_2=\{X_3\}$. Then, the reduced potential $\Gamma$ is the restriction of $\gamma$ to those rows in the table where $X_1=X_2$. Hence $\Gamma$ has four rows in its table and equivalently can be thought of defining a potential over the $4$ possible combinations of $Y_1$ and $Y_2$ variables. We are now ready to define a reduced graphical model.

\begin{defn}
Let $G=(\mathcal{X},\gamma)$ represent a PGM. Given a partition $\mathcal{Y}^P$ of $\mathcal{X}$, the reduced graphical model $\mathcal{G}(\mathcal{Y},\Gamma)$ is the graphical model defined over the set of partition variables $\mathcal{Y}$ such that every potential $\gamma_j \in \gamma$ in $G$ is replaced by the corresponding reduced potential $\Gamma_j \in \Gamma$ in $\mathcal{G}$.
\end{defn}
Let $E({\bf x})$ and $\mathcal{E}({\bf y})$ denote the energies of the states ${\bf x}$ and ${\bf y}$ in $G$ and $\mathcal{G}$, respectively. The following theorem relates the energies of the states in the two graphical models.

\begin{thm}\label{thm:energy_correspondence}
For every assignment ${\bf y}$ of $\mathcal{Y}$ in $\mathcal{G}$, there is a corresponding assignment ${\bf x}$ of $\mathcal{X}$ such that $\mathcal{E}({\bf y}) = E({\bf x})$.
\end{thm}
The theorem can be proved by noting that each potential $\Gamma_j(\hat{Y}_j)$ in $\mathcal{G}$ was obtained by restricting the original potential $\gamma_j(\hat{X}_j)$ to those assignments where variables in $X_j$ belonging to the same partition took the same label. Since this correspondence is true for every potential in the reduced set, to obtain the desired state ${\bf x}$, for every variable $X_i \in {\mathcal X}$ we simply assign it the label of its partition in ${\bf y}$. 
\begin{cor}\label{cor:map_energy_correspondence}
Let ${\bf x}_{\rm MAP}$ and ${\bf y}_{\rm MAP}$ be the MAP states (i.e. having the minimum energy) for $G$ and $\mathcal{G}$, respectively. Then, $\mathcal{E}({\bf y}_{\rm MAP}) \ge E({\bf x}_{\rm MAP})$.
\end{cor}
The process of reduction can be seen as curtailing the entire search space to those assignments where variables in the same partition take the same label. A reduction in the problem space will lead to computational gains but might result in loss of solution quality, where the solution quality can be captured by the difference between $\mathcal{E}({\bf y}_{\rm MAP})$ and $E({\bf x}_{\rm MAP})$. Therefore, we need to trade-off the balance between the two.

Intuitively, a good problem reduction will keep those variables in the same partition which are likely to have the same value in the optimal assignment for the original problem. How do we find such variables without actually solving the inference task? We will describe one such technique in Section~\ref{sec:color_passing_vision}. 

There is another perspective. Instead of solving one reduced problem, we can instead work with a series of reduced problems which successively get closer to the optimal solution. The initial reductions are coarser and far from optimal, but can be solved efficiently to quickly reach in the region where the solution lies. The successive iterations can then refine the solution iteratively getting closer to the optimal. This leads us to the coarse-to-fine inference described next.

\subsection{Coarse to Fine Inference}
We will define a framework for C2F (coarse-to-fine) inference so that we maintain the computational advantage while still preserving optimality. In the following, for ease of notation, we will drop the superscript $P$ in $\mathcal{Y}^p$ to denote the partition of $\mathcal{X}$. Therefore, $\mathcal{Y}$ will refer to both the partition as well as the set of partition variables.
Before we describe our algorithm, let us start with some definitions.
\begin{defn}\label{defn:coarse_partn}
Let $\mathcal{Y}$ and $\mathcal{Y'}$ be two partitions of $\mathcal{X}$. We say that $\mathcal{Y}$ is coarser than $\mathcal{Y}'$, denoted as $\mathcal{Y} \preceq \mathcal{Y}'$, if $\forall y' \in \mathcal{Y}' \exists y \in \mathcal{Y}$ such that $y' \subseteq y$. We equivalently say that $\mathcal{Y}'$ is finer than $\mathcal{Y}$.
\end{defn}
It is easy to see that $\mathcal{X}$ defines a partition of itself which is the finest among all partitions, i.e., $\forall \mathcal{Y}$ such that ${\mathcal Y}$ is a partition of $\mathcal{X}$, $\mathcal{Y} \preceq \mathcal{X}$. We also refer it to as the degenerate partition. For ease of notation, we will denote the finest partition by $\mathcal{Y}^{*}$ (same as $\mathcal{X}$). We will refer to the corresponding PGM as $\mathcal{G}^{*}$ (same as $G$). Next, we state a theorem which relates two partitions with each other.
\begin{lemma}\label{thm:finer_partitions}
Let $\mathcal{Y}$ and $\mathcal{Y}'$ be two partitions of $\mathcal{X}$ such that $\mathcal{Y} \preceq \mathcal{Y}'$. Then $\mathcal{Y}'$ can be seen as a partition of the set $\mathcal{Y}$. 
\end{lemma}
The proof of this lemma is straightforward and is omitted due to lack of space. 
Consider a set ${\mathcal{\bf Y}}$ of coarse to fine partitions given as $\mathcal{Y}^0 \preceq \mathcal{Y}^1,\cdots,\preceq,\mathcal{Y}^t,\preceq,\cdots,\mathcal{Y}^*$. Let $\mathcal{G}^t,\mathcal{E}^t,{\bf y}_{MAP}^{t}$ respectively denote the reduced problem, energy function and MAP assignment for the partition $\mathcal{Y}^t$. Using Lemma~\ref{thm:finer_partitions}, $\mathcal{Y}^{t+1}$ is a partition of $\mathcal{Y}^{t}$. Then, using Theorem~\ref{thm:energy_correspondence}, we have for every assignment ${\bf y^{t}}$ to variables in $\mathcal{Y}^t$, there is an assignment ${\bf y^{t+1}}$ to variables in $\mathcal{Y}^{t+1}$ such that $\mathcal{E}^{t}({\bf y^t})=\mathcal{E}^{t+1}({\bf y^{t+1}})$. Also, using Corollary~\ref{cor:map_energy_correspondence}, we have $\forall t\mathrm{\ }\mathcal{E}^{t}({\bf y^{t}_{\rm MAP}}) \ge \mathcal{E}^{t+1}({\bf y^{t+1}_{\rm MAP}})$. Together, these two statements imply that starting from the coarsest partition, we can gradually keep on improving the solution as we move to finer partitions.

Our C2F set-up assumes an iterative MAP inference algorithm $A$ which has the anytime property i.e., can produce solutions of increasing quality with time.
C2F Function (see Algorithm 1) takes 3 inputs: a set of C2F partitions $\mathcal{\bf Y}$, inference algorithm $A$, and a stopping criteria $\mathcal{C}$. The algorithm $A$ in turn takes three inputs: PGM $\mathcal{G}^t$, starting\_assignment ${\mathcal {\bf y}^t}$, stopping criteria $\mathcal{C}$. $A$ outputs an approximation to the MAP solution once the stopping criteria $\mathcal{C}$ is met. Starting with the coarsest partition ($t=0$ in line 2), a start state is picked for the coarsest problem to be solved (line 3). In each iteration (line 4), C2F finds the MAP estimate for the current problem ($\mathcal{G}^t$) using algorithm $A$ (line 5). This solution is then mapped to a same energy solution of the next finer partition (line 6) which becomes the starting state for the next run of A. The solution is thus successively refined in each iteration. The process is repeated until we reach the finest level of partition. In the end, $A$ is run on the finest partition and the resultant solution is output (lines 9,10). Since the last partition in the set is the original problem $\mathcal{G}^{*}$, optimality with respect to $A$ is guaranteed. 

Next, we describe how to use the color passing algorithm (Section~\ref{sec:background}) to get a series of partitions which get successively finer. Our C2F algorithm can then be applied on this set of partitions to get anytime solutions of high quality while being computationally efficient.

{\small
\begin{algorithm}\label{algo:C2F}
\caption{Coarse-to-Fine Lifted MAP Algorithm}
\begin{algorithmic}[1]
\State \hspace{-0.15in}{\bf C2F\_Lifted\_MAP}{(C2F Partitions ${\mathcal{\bf Y}}$, Algo $A$,Criteria $\mathcal{C}$)}
    \State $t=0$; $T=|\mathcal{\bf Y}|$;
    \State ${\bf y}^{t}$ = getInitState($\mathcal{G}^{t}$);
    \State {\bf While} ($t < T$);
         \State \hspace{0.15in} ${\bf y}^{t}_{\rm MAP}$ = $A(\mathcal{G}^t,y^{t},\mathcal{C})$;
         \State \hspace{0.15in} ${\bf y}^{t+1} = getEquivAssignment(\mathcal{Y}^{t},\mathcal{Y}^{t+1},{\bf y}^{t}_{\rm MAP})$;
         \State \hspace{0.15in} $t = t+1$;
    \State {\bf END\_While}
    \State ${\bf y}^{T}_{\rm MAP}$ = $A(\mathcal{G}^T,y^{T},\mathcal{C})$;
    \State {\bf return} $y^{T}_{\rm MAP}$
\end{algorithmic}
\end{algorithm}
}

\subsection{C2F Partitioning for Computer Vision}\label{sec:color_passing_vision} \label{sec:c2f4cv}
We now adapt the general color passing algorithm to MRFs for CV problems. Unfortunately, unary potentials make color passing highly ineffective. Different pixels have different RGB values and intensities, leading to almost every pixel getting a different unary potential. Naive application of color passing splits almost all variables into their own partitions, and lifting offers little value.

A natural approximation is to define a threshold, such that two unary potentials within that threshold be initialized with the same color. Our experiments show limited success with this scheme because because two pixels may have the same label even when their actual unary potentials are very different. What is more
important is relative importance given to each label than the actual potential value.

In response, we adapt color passing for CV by initializing it as before, but with one key change: we initialize two unary potential nodes with the same color if their lowest energy labels have the same order for the top $N_L$ labels (we call this unary split threshold). Experiments reveal that this approximation leads to effective partitions for lifted inference.

Finally, we can easily construct a sequence of coarse-to-fine partitions in the natural course of color passing's execution -- every iteration of color passing creates a finer partition. Moreover, as an alternative approach, we may also increase $N_L$. In our implementations, we intersperse the two, i.e., before every next step we pick one of two choices: either, we run another iteration of color passing; or, we increase $N_L$ by one, and split each variable partition based on the $N_L^\mathrm{th}$ lowest energy labels of its constituent variables.

We parameterize $CP(N_L, N_{iter})$ to denote the partition from the current state of color passing, which has been run till $N_{iter}$ iterations and unary split threshold is $N_L$. It is easy to prove that another iteration of color passing or splitting by increasing $N_L$ as above leads to a finer partition. I.e., $CP(N_{L}, N_{iter}) \preceq CP(N_{L}+1, N_{iter})$ and $CP(N_{L}, N_{iter}) \preceq CP(N_{L}, N_{iter}+1)$. We refer to each element of a partition of variables as a {\em lifted pixel}, since it is a subset of pixels.

\begin{figure*}
\centering
\framebox{
\subfigure[]{
{\includegraphics[width=0.25\textwidth]{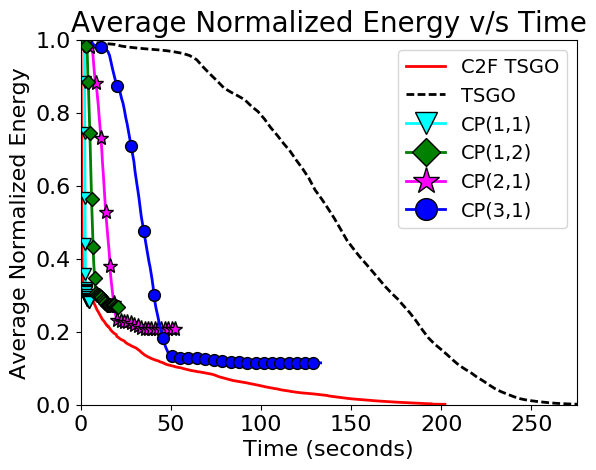}}
\label{fig:stereo_energy}}
\subfigure[]{
{\includegraphics[width=0.25\textwidth]{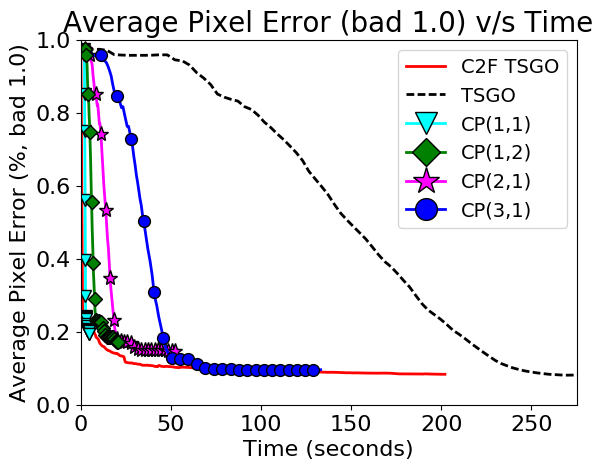}}
\label{fig:stereo_error}}
}
\hspace{0.5cm}
\subfigure[]{
{\includegraphics[width=0.25\textwidth]{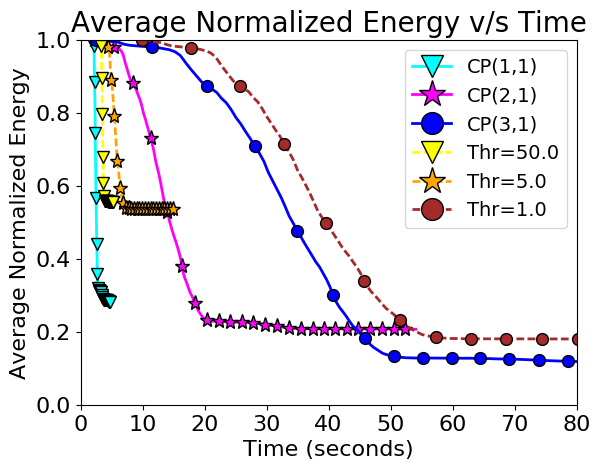}}
\label{fig:threshold_energy}}
\caption {\small {{\bf (a)} Average (normalized) energy vs. inference time {\bf (b)} Average pixel error vs. time. C2F TSGO achieves roughly 60\% reduction in time for reaching the optima. It has best anytime performance compared to vanilla TSGO and static lifted versions. {\bf (c)} Average (normalized) energy vs. time for different thresholding values and CP partitions. Plots with the same marker have MRFs of similar sizes. 
}}
\label{fig:tsgoplots}

\end{figure*}

\begin{figure*}
\centering
\subfigure[]{
{\includegraphics[width=0.15\textwidth]{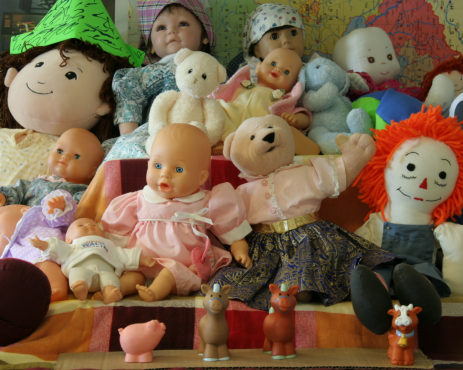}}
{\includegraphics[width=0.15\textwidth]{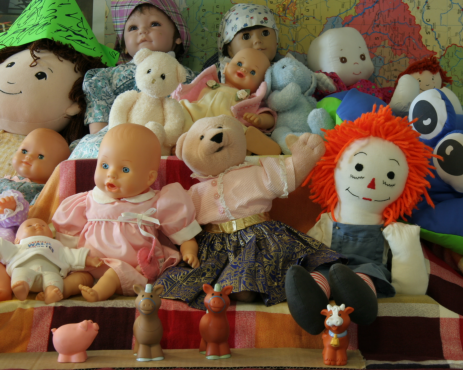}}\label{fig:original}}
\subfigure[]{
{\includegraphics[width=0.15\textwidth]{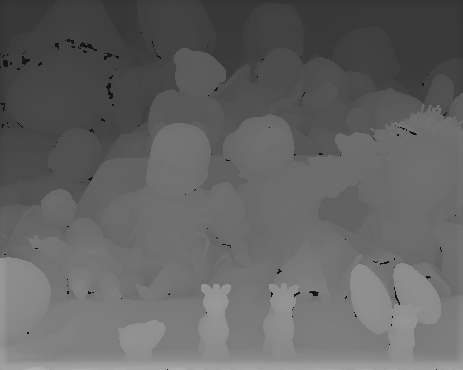}}\label{fig:gold}}
\subfigure[]{
{\includegraphics[width=0.15\textwidth]{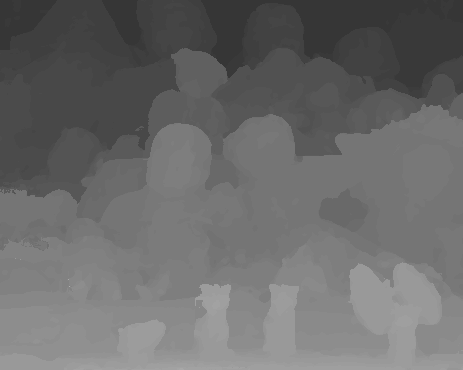}}\label{fig:tsgo}}
\subfigure[]{
{\includegraphics[width=0.15\textwidth]{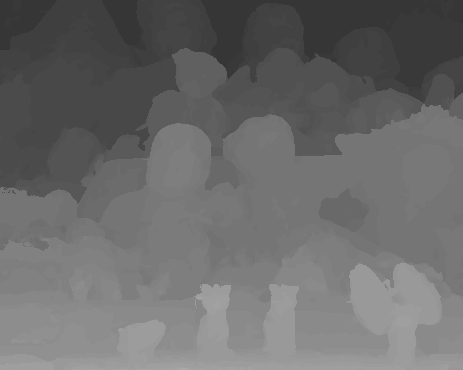}}\label{fig:C2F}}
\subfigure[]{
{\includegraphics[width=0.15\textwidth]{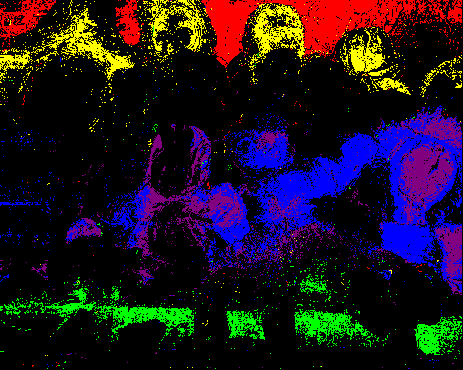}}\label{fig:sym_map_stereo}}
\vspace*{-2ex}
\caption{\small{Qualitative results for Doll image at convergence. C2F-TSGO is similar to base TSGO.{\bf (a)} Left and Right Images {\bf (b)} Ground Truth {\bf (c)} Disparity Map by TSGO {\bf (d)} Disparity Map by C2F TSGO {\bf (e)} Each colored region (other than black) is one among the 10 largest partition elements from CP(1,1). Each color represents one partition element. Partition elements form non-contiguous regions}}
\label{fig:stereoQualitative}
\vspace*{-1ex}
\end{figure*}

\section{Lifted Inference for Stereo Matching} \label{sec:stereo} 

We first demonstrate the value of lifted inference in the context of stereo matching \cite{Scharstein2002}. It aims to find pixel correspondences in a set of images of the same scene, which can be used to further estimate the 3D scene. Formally, two images \(I^l\) and \(I^r\) corresponding to images of the scene from a left camera and a right camera are taken such that both cameras are at same horizontal level. The goal is to compute a disparity labeling 
\(D^l\) for every pixel \(X = (a,b)\) such that \(I^l[a][b]\) corresponds to \(I^r[a - D^l[a][b]][b]\).
We build a lifted version of TSGO \cite{TSGO}, as it is MRF-based and ranks $2^{nd}$ on the Middlebury Stereo Evaluation Version 2 leaderboard.\footnote{\em http://vision.middlebury.edu/stereo/eval/}

\vspace{0.5ex}
\noindent{\bf Background on TSGO: }
TSGO treats stereo matching as a two-step energy minimization, where the first step is on a fully connected MRF with pairwise potentials and the second is on a conventional locally connected MRF. Lack of space precludes a detailed description of the first step. At the high level, TSGO runs one iteration of message passing on fully connected MRF, computes marginals of each pixel $X$, which act as unary potentials $\phi(X)$ for the MRF of second step.   

The pairwise potential $\psi$ used in step two is $\psi(X, X') =w(X,X')\varphi(X,X')$, where 
\(\varphi(X,X')\) is a truncated linear function of \(\Vert X-X' \Vert\), and \({w}(X,X')\) takes one of three distinct values depending on color difference between pixels. The MAP assignment $\mathbf{x}_\mathrm{MAP}$ computes the lowest energy assignment of disparities $D^l$ for every pixel for this MRF.

\vspace{0.5ex}
\noindent{\bf Lifted TSGO: } Since step two is costlier, we build its lifted version as discussed in previous section. For color passing, two unary potential nodes are initialized with the same color if their lowest energy labels exactly match ($N_L=1$). Other initializations are consistent with original color passing for general MRFs. 
A sequence of coarse-to-fine models is outputted as per Section \ref{sec:c2f4cv}. C2F TSGO uses outputs from the sequence $CP(1,1)$, $CP(2,1)$, $CP(3,1)$ and then refines to the original MRF. Model refinement is triggered whenever energy hasn't decreased in the last four iterations of alpha expansion (this becomes the stopping criteria $\mathcal{C}$ in Algorithm 1).

\vspace{0.5ex}
\noindent{\bf Experiments: }
Our experiments build on top of the existing TSGO implementation\footnote{\em  http://www.cvc.uab.es/$\sim$mozerov/Stereo/}, but we  change the minimization algorithm in step two to alpha expansion fusion \cite{5166447} from OpenGM2 library \cite{opengm1,kappes2013comparative}, as it improves the speed of the base implementation. 
We use the benchmark Middlebury Stereo datasets of 2003, 2005 and 2006 \cite{Scharstein:2003:HSD:1965841.1965865,Hirschmuller2007}. For the 2003 dataset, quarter-size images are used and for others, third-size images are used. The label space is of size 85 (85 distinct disparity labels).

We compare our coarse-to-fine TSGO (using $CP(N_L,N_{iter})$ partitions) against vanilla TSGO.
Figures \ref{fig:tsgoplots}(a,b) show the aggregate plots of energy (and error) vs. time. We observe that C2F TSGO reaches the same optima as TSGO, but in less than half the time. It has a much superior anytime performance -- if inference time is given as a deadline, C2F TSGO obtains 59.59\% less error on average over randomly sampled deadlines.
We also eyeball the outputs of C2F TSGO and TSGO and find them to be visually similar. Figure \ref{fig:stereoQualitative} shows a sample qualitative comparison. 
Figure \ref{fig:sym_map_stereo} shows five of the ten largest partition elements in the partition from $CP(1,1)$. Clearly, the partition elements formed are not contiguous, and seem to capture variables that are likely to get the same assignment. This underscores the value of our lifting framework for CV problems.

We also compare our $CP(N_L,N_{iter})$ partitioning strategy with threshold partitioning discussed in Section \ref{sec:c2f4cv}. We merge two pixels in thresholding scheme if the L1-norm distance of their unary potentials is less than a threshold. For each partition induced by our approach, we find a value of threshold that has roughly the same number of lifted pixels. Figure \ref{fig:tsgoplots}(c) shows that partitions based on $CP(1,1)$ and $CP(3,1)$ converges to a much lower energy quickly compared to the corresponding threshold values ($Thr=50$ and $Thr=1$ respectively). For $CP(2,1)$, convergence is slower compared to corresponding threshold ($Thr=5$) but eventually $CP(2,1)$ has significantly better quality. 

\begin{figure*} 
\centering
\subfigure{
{\includegraphics[width=0.2\textwidth]{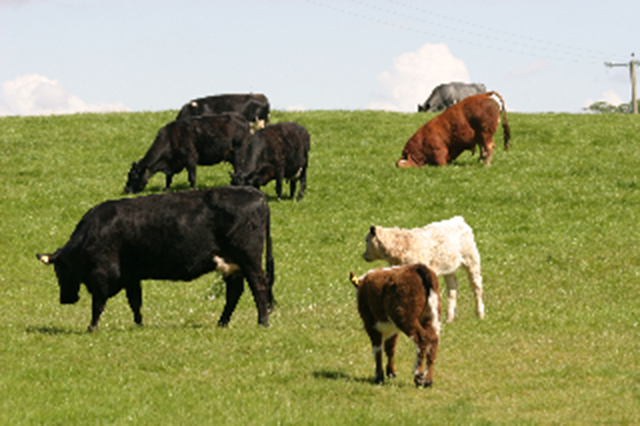}}}
\hspace*{1ex}
\subfigure{
{\includegraphics[width=0.2\textwidth]{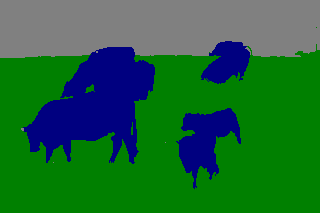}}}
\subfigure{
{\includegraphics[width=0.2\textwidth]{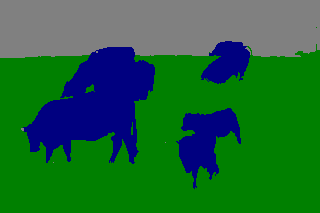}}}
\hspace*{1ex}
\subfigure{
{\includegraphics[width=0.30\textwidth]{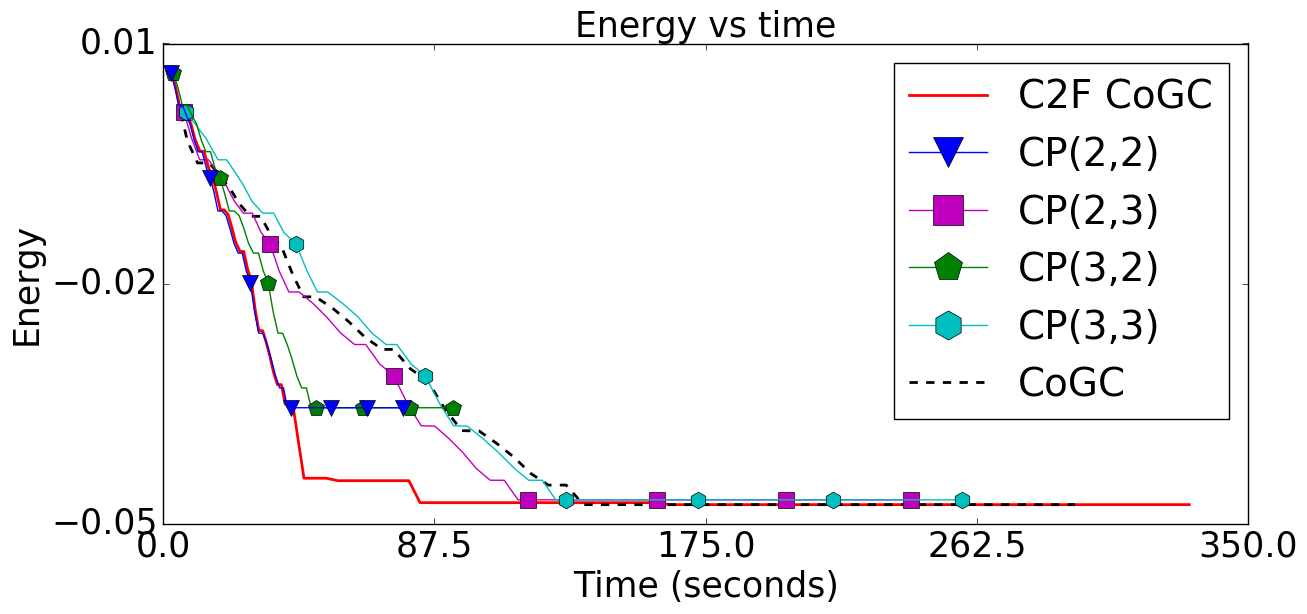}}}

\centering
\subfigure{
{\includegraphics[width=0.2\textwidth]{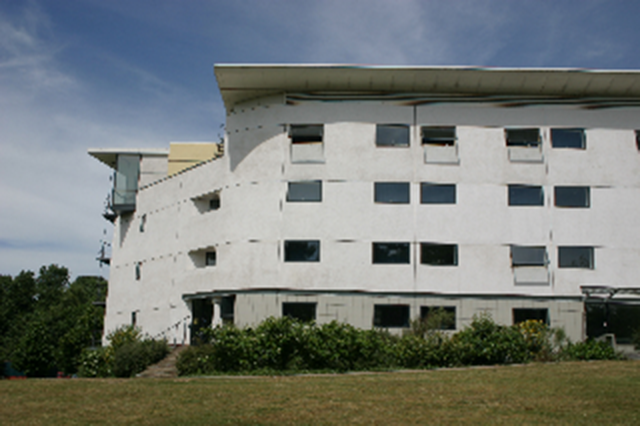}}}
\hspace*{1ex}
\subfigure{
{\includegraphics[width=0.2\textwidth]{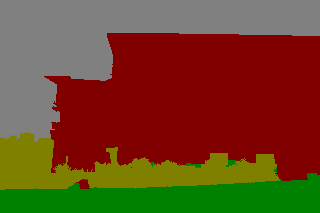}}}
\subfigure{
{\includegraphics[width=0.2\textwidth]{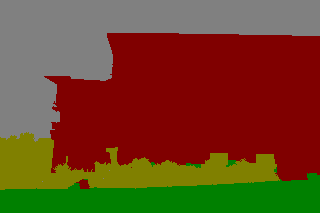}}}
\hspace*{1ex}
\subfigure{
{\includegraphics[width=0.3\textwidth]{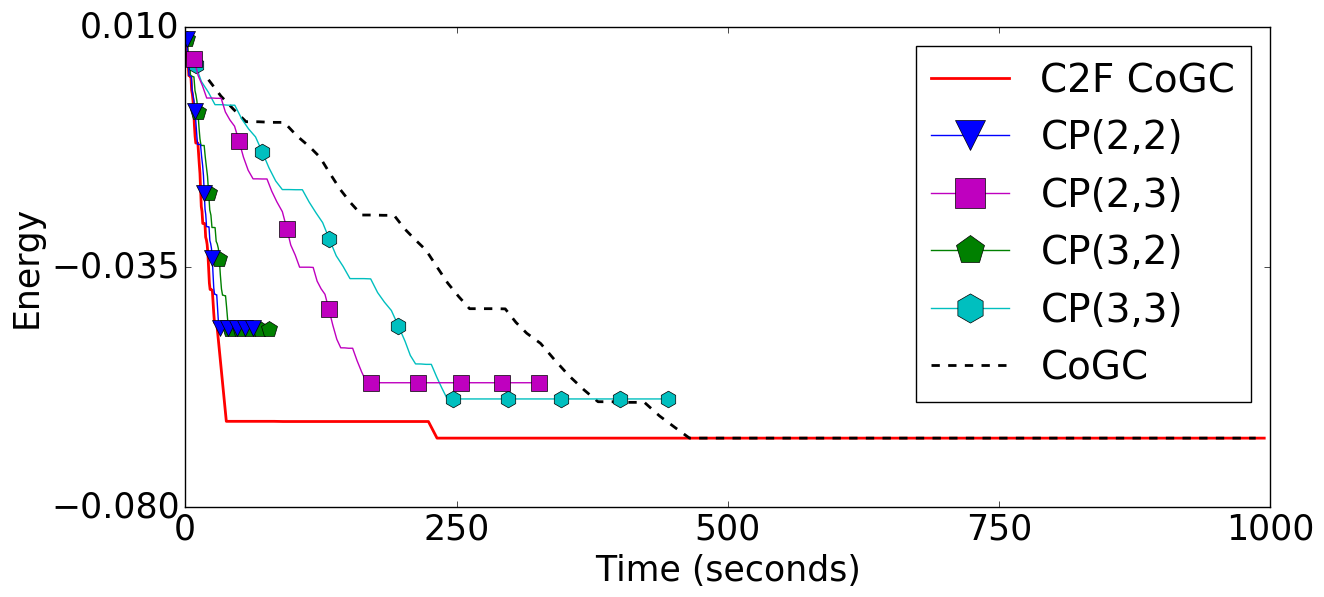}}}

\centering
\subfigure{
\includegraphics[width=0.2\textwidth]{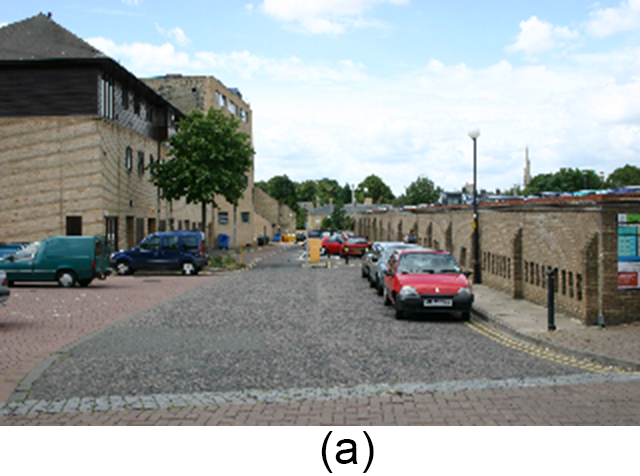}
}
\hspace*{1ex}
\subfigure{
{\includegraphics[width=0.2\textwidth]{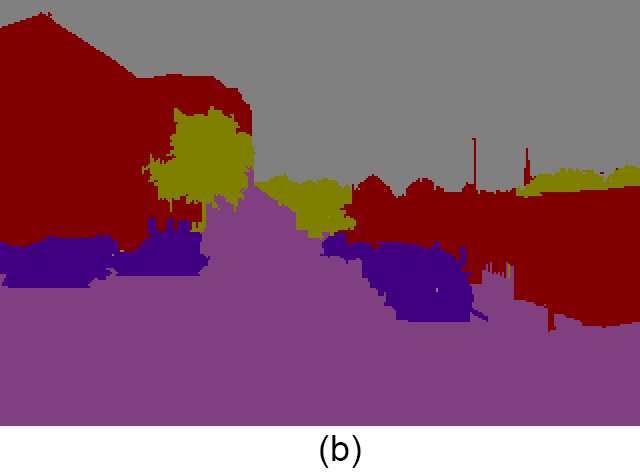}}
}
\subfigure{
{\includegraphics[width=0.2\textwidth]{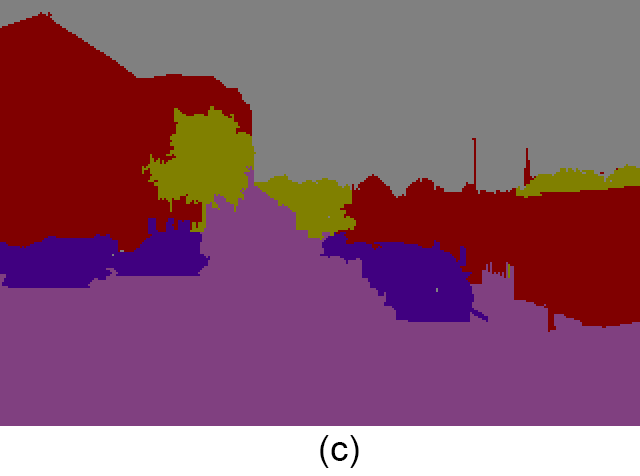}}
}
\hspace*{1ex}
\subfigure{
{\includegraphics[width=0.3\textwidth]{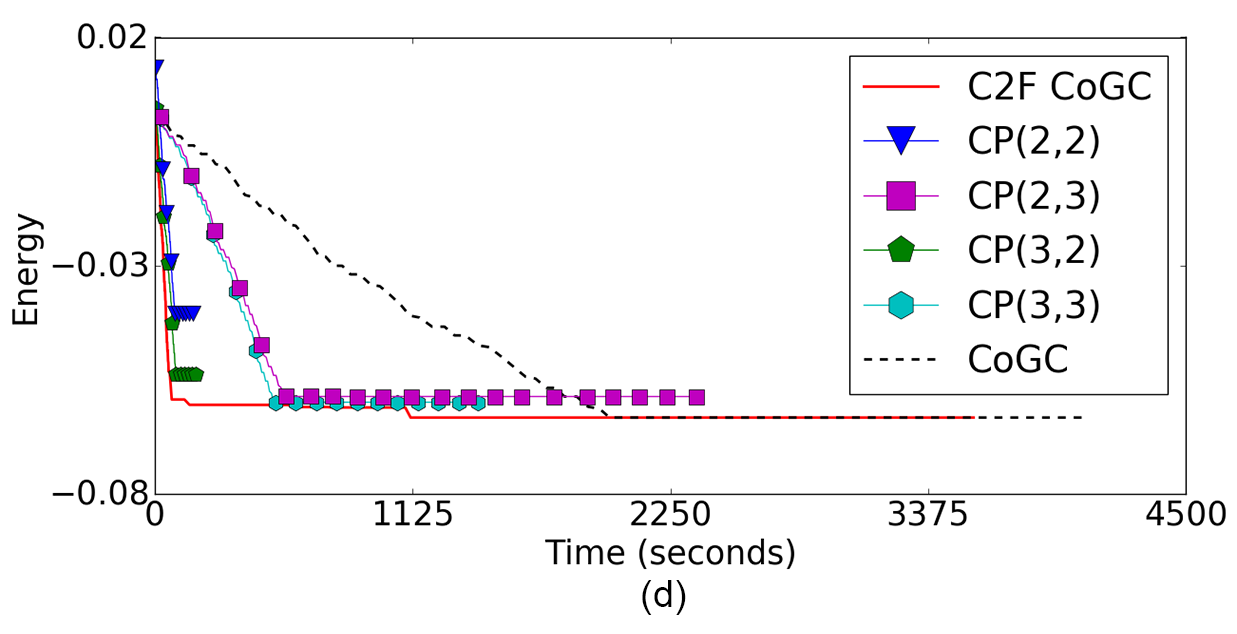}}}
\caption{\small {{\bf (a-c)} Qualitative Results for Segmentation. C2F has quality similar to CoGC algorithm {\bf (a)} Original Image {\bf (b)} Segmentation by CoGC {\bf (c)} Segmentation by C2F CoGC {\bf (d)} C2F CoGC has lower energy compared to CoGC and other lifted variant at all times}}
\label{fig:segAll}
\vspace*{-2ex}
\end{figure*}

\section{Lifted Inference for Image Segmentation} \label{sec:segmentation}
We now demonstrate the general nature of our lifted CV framework by applying it to a second task. We choose multi-label interactive image segmentation, where the goal is to segment an image \(I\) based on a seed labeling (true labels for a few pixels) provided as input. Like many other CV problems, this also has an MRF-based solution, with the best label-assignment generally obtained by MAP inference using graph cuts or loopy belief propagation  \cite{boykov2001fast,szeliski&al06}.

However, MRFs with only pairwise potentials are known to suffer from {\em short-boundary bias}  -- they prefer segmentations with shorter boundaries, because pairwise potentials penalize every pair of boundary pixels. This leads to incorrect labeling for sharp edge objects. Kohli \etal\ \shortcite{kohli2013principled} use CoGC, cooperative graph cuts \cite{jegelka11}, to develop one of the best MRF-based solvers that overcome this bias.

\vspace{0.5ex}
\noindent{\bf Background on CoGC: } Traditional MRFs linearly penalize the number of label discontinuities at edges (boundary pixel pairs), but CoGC penalizes the number of {\em types} of label discontinuities through the use of a concave energy function over groups of ordered edges. It first clusters all edges on the basis of color differences, and later applies a concave function separately over the number of times a specific discontinuity type is present in each edge group $g\in \mathbb{G}$. Their carefully engineered CoGC energy function is as follows:
{\small
\[
E(\mathrm{\textbf{x}}) = \sum_{i=1}^{|{\cal X}|}  \phi_{i}(x_i) + \sum_{g\in \mathbb{G}}\sum_{l\in L}F\left(\sum_{(x,x')\in g}w(x,x').\mathbb{I}(x=l, x'\neq l) \right)
\]
}
\noindent where unary potentials $\phi$ depend on colors of seed pixels, $F$ is a concave function, $\mathbb{I}$ the indicator function, and $w(x,x')$ depends on the color difference between $x,x'$. Intuitively, $F$ collects all edges with similar discontinuities and penalizes them sub-linearly, thus reducing the short-boundary bias in the model.
The usage of a concave function makes the MRF higher order with cliques over edge groups. However, the model is shown to reduce to a pairwise hierarchical MRF through the addition of auxiliary variables.\\
\vspace{0.5ex}
\noindent{\bf Lifted CoGC: }
CoGC is lifted using the framework of Section \ref{sec:framework}, with one additional change. We cluster edge groups using color difference {\em and} the position of the edge. Edge groups that are formed only on the basis of color difference make the error of grouping different segment's boundaries into a single group. For e.g., it erroneously cluster boundaries between white cow and grass, and sky and grass together in the top image in Figure \ref{fig:segAll}.

Coarse-to-fine partitions are obtained by the method described in Section \ref{sec:c2f4cv}. C2F CoGC uses outputs from the sequence \(CP(\lceil\frac{ L }{2}\rceil, 2), CP(\lceil\frac{ L }{2}\rceil, 3)\) before refining to the original MRF. Model refinement is triggered if energy has not reduced over the last \(\vert L \vert\) iterations.

\vspace{0.5ex}
\noindent{\bf Experiments: }
Our experiments use the implementation of Cooperative Graph Cuts as provided by \cite{kohli2013principled}.\footnote{Available at {\em  https://github.com/aosokin/coopCuts\_CVPR2013}} Energy minimization is performed using alpha expansion \cite{boykov2001fast}. The implementation of CoGC performs a greedy descent on auxiliary variables while performing alpha expansion on the remaining variables, as described in Kohli et. al. \shortcite{kohli2013principled}.
The dataset used is provided with the implementation. It is a part of the MSRC V2 dataset.\footnote{Available at \emph{ https://www.microsoft.com/en-us/research/ project/image-understanding/?from=http\%3A\%2F\%2Fresearch. microsoft.com\%2Fvision\%2Fcambridge\%2Frecognition\%2F}}.

Figure \ref{fig:segAll} shows three individual energy vs. time plots. Results on other images are similar. We find that C2F CoGC algorithm converges to the same energy as CoGC in about two-thirds the time on average. Overall, C2F CoGC achieves a much better anytime performance than other lifted and unlifted CoGC. 

Similar to Section \ref{sec:stereo}, refined partitions attain better quality than coarser ones at the expense of time. Since the implementation performs a greedy descent over auxiliary variables, refinement of current partition also resets the auxiliary variables to the last value that produced a change. Notice that energy minimization on output of \(CP(2,3)\) attains a lower energy than on \(CP(3,2)\). This observation drives our decision to refine by increasing \(N_{iter}\). Qualitatively, C2F CoGC produces the same labeling as CoGC. Finally, similar to stereo matching, partitions based on thresholding scheme perform significantly worse compared to $CP(N_L,N_{iter})$ for image segmentation as well.

\section{Related Work}
There is a large body of work on exact lifting, both marginal~\cite{kersting&al09,gogate&domingos11,niepert12,mittal&al15} and MAP~\cite{kersting&al09,gogate&domingos11,niepert12,sarkhel&al14,mittal&al14}, which is not directly applicable to our setting. There is some recent work on approximate lifting~\cite{broeck&darwiche13,venugopal&gogate14evidence,singla&al14,sarkhel&al15,broeck&niepert15} but it's focus is on marginal inference whereas we are interested in lifted MAP. Further, this work can't handle a distinct unary potential on every node. An exception is work by Bui et al.~\shortcite{bui&al12} which explicitly deals with lifting in presence of distinct unary potentials. Unfortunately, they make a very strong assumption of exchangeability in the absence of unaries which does not hold true in our setting since each pixel has its own unique neighborhood.

Work by Sarkhel et al.~\shortcite{sarkhel&al15} is probably the closest to our work. They design a C2F hierarchy to cluster constants for approximate lifted MAP inference in Markov logic. In contrast, we partition ground atoms in a PGM. Like other work on approximate lifting, they can't handle distinct unary potentials. Furthermore, they assume that their theory is provided in a normal form, i.e., without evidence, which can be a severe restriction for most practical applications. Kiddon \& Domingos~\shortcite{kiddon&domingos11} also propose C2F inference for an underlying Markov logic theory. They use a hierarchy of partitions based on a pre-specified ontology. CV does not have any such ontology available, and needs to discover partitions using the PGM directly.

Nath \& Domingos~\shortcite{nath&domingos10} exploit (approximate) lifted inference for video segmentation. They experiment on a specific video problem (different from ours), and they only compare against vanilla BP. Their initial partitioning scheme is similar to our thresholding approach, which does not work well in our experiments.

In computer vision, a popular approach to reduce the complexity of inference is to use superpixels \cite{achanta2012,bergh&al12}. Superpixels are obtained by merging neighboring nodes that have similar characteristics. All pixel nodes in the same superpixel are assigned the same value during MAP inference. SLIC \cite{achanta2012} is one of the most popular algorithms for discovering superpixels. Our approach differs from SLIC in some significant ways. First, their superpixels are local in nature whereas our algorithm can merge pixels that are far apart. This can help in merging two disconnected regions of the same object in a single lifted pixel. Second, they obtain superpixels independent of the inference algorithm, whereas we tightly integrate our lifting with the underlying inference algorithm. This can potentially lead to discovery of better partitions; indeed, this helped us tremendously in image segmentation. Third, they do not provide a C2F version of their algorithm and we did not find it straightforward to extend their approach to discover successively finer partitions. There is some recent work ~\cite{WeiXingSH} which addresses last two of these challenges by introducing a hierarchy of superpixels. In our preliminary experiments, we found that SLIC and superpixel hierarchy perform worse than our lifting approach. Performing more rigorous comparisons is a direction for future work.

\section{Conclusion and Future Work}
We develop a generic template for applying lifted inference to structured output prediction tasks in computer vision. We show that MRF-based CV algorithms can be lifted at different levels of abstraction, leading to methods for coarse to fine inference over a sequence of lifted models. We test our ideas on two different CV tasks of stereo matching and interactive image segmentation. We find that C2F lifting is vastly more efficient than unlifted algorithms on both tasks obtaining a superior anytime performance, and without any loss in final solution quality. To the best of our knowledge, this is the first demonstration of lifted inference in conjunction with top of the line task-specific algorithms. Although we restrict to CV in this work, we believe that our ideas are general and can be adapted to other domains such as NLP, and computational biology. We plan to explore this in the future.

\section*{Acknowledgements}

We thank anonymous reviewers for their comments and suggestions. Ankit Anand is being supported by the TCS Research Scholars Program. Mausam is being supported by grants from Google and Bloomberg. Parag Singla is being supported by a DARPA grant funded under the Explainable AI (XAI) program. Both Mausam and Parag Singla are being supported by the Visvesvaraya Young Faculty Fellowships by Govt. of India. Any opinions, findings, conclusions or recommendations expressed in this paper are those of the authors and do not necessarily reflect the views or official policies, either expressed or implied, of the funding agencies.
{\small
\bibliographystyle{named}
\bibliography{all}

\begin{thebibliography}{}

\bibitem[\protect\citeauthoryear{Achanta \bgroup \em et al.\egroup
  }{2012}]{achanta2012}
R.~Achanta, A.~Shaji, K.~Smith, A.~Lucchi, P.~Fua, and S.~Süsstrunk.
\newblock {SLIC Superpixels Compared to State-of-the-Art Superpixel Methods}.
\newblock {\em In PAMI}, Nov 2012.

\bibitem[\protect\citeauthoryear{Anand \bgroup \em et al.\egroup
  }{2016}]{anand&al16}
A.~Anand, A.~Grover, Mausam, and P.~Singla.
\newblock {Contextual Symmetries in Probabilistic Graphical Models}.
\newblock In {\em IJCAI}, 2016.

\bibitem[\protect\citeauthoryear{Anand \bgroup \em et al.\egroup
  }{2017}]{anand&al17}
A.~Anand, R.~Noothigattu, P.~Singla, and Mausam.
\newblock {Non-Count Symmetries in Boolean \& Multi-Valued Prob. Graphical
  Models}.
\newblock In {\em AISTATS}, 2017.

\bibitem[\protect\citeauthoryear{Andres \bgroup \em et al.\egroup
  }{2010}]{opengm1}
B.~Andres, J.~H. Kappes, U.~K\"{o}the, C.~Schn\"{o}rr, and F.~A. Hamprecht.
\newblock {An Empirical Comparison of Inference Algorithms for Graphical Models
  with Higher Order Factors Using OpenGM}.
\newblock In {\em Pattern Recognition}. 2010.

\bibitem[\protect\citeauthoryear{Blei \bgroup \em et al.\egroup
  }{2003}]{blei&al03}
D.~Blei, A.~Ng, and M.~Jordan.
\newblock {Latent Dirichlet Allocation}.
\newblock {\em JMLR}, 3, March 2003.

\bibitem[\protect\citeauthoryear{Boykov \bgroup \em et al.\egroup
  }{2001}]{boykov2001fast}
Y.~Boykov, O.~Veksler, and R.~Zabih.
\newblock {Fast Approximate Energy Minimization via Graph Cuts}.
\newblock {\em In PAMI}, 23(11), November 2001.

\bibitem[\protect\citeauthoryear{Braz \bgroup \em et al.\egroup
  }{2005}]{braz&al05}
R.~Braz, E.~Amir, and D.~Roth.
\newblock {Lifted First-Order Probabilistic Inference}.
\newblock In {\em IJCAI}, 2005.

\bibitem[\protect\citeauthoryear{Bui \bgroup \em et al.\egroup
  }{2012}]{bui&al12}
H.~Bui, T.~Huynh, and R.~De~Salvo~Braz.
\newblock {Exact Lifted Inference with Distinct Soft Evidence on Every Object}.
\newblock In {\em AAAI}, 2012.

\bibitem[\protect\citeauthoryear{Bui \bgroup \em et al.\egroup
  }{2013}]{bui&al13}
H.~Bui, T.~Huynh, and S.~Riedel.
\newblock {Automorphism Groups of Graphical Models and Lifted Variational
  Inference}.
\newblock In {\em UAI}, 2013.

\bibitem[\protect\citeauthoryear{Freeman \bgroup \em et al.\egroup
  }{2000}]{freeman&al00}
W.~Freeman, E.~Pasztor, and O.~Carmichael.
\newblock {Learning Low-Level Vision}.
\newblock {\em In IJCV}, 40, 2000.

\bibitem[\protect\citeauthoryear{Friedman}{2004}]{friedman04}
N.~Friedman.
\newblock {Inferring Cellular Networks using Probabilistic Graphical Models}.
\newblock {\em Science}, 303, 2004.

\bibitem[\protect\citeauthoryear{Gogate and Domingos}{2011}]{gogate&domingos11}
V.~Gogate and P.~Domingos.
\newblock {Probabilisitic Theorem Proving}.
\newblock In {\em UAI}, 2011.

\bibitem[\protect\citeauthoryear{Hirschmuller and
  Scharstein}{2007}]{Hirschmuller2007}
H.~Hirschmuller and D.~Scharstein.
\newblock {Evaluation of Cost Functions for Stereo Matching}.
\newblock In {\em CVPR}, 2007.

\bibitem[\protect\citeauthoryear{Jegelka and Bilmes}{2011}]{jegelka11}
S.~Jegelka and J.~Bilmes.
\newblock {Submodularity Beyond Submodular Energies: Coupling Edges in Graph
  Cuts}.
\newblock In {\em CVPR}, 2011.

\bibitem[\protect\citeauthoryear{Jernite \bgroup \em et al.\egroup
  }{2015}]{jernite&al15}
Y.~Jernite, A.~Rush, and D.~Sontag.
\newblock {A Fast Variational Approach for Learning Markov Random Field
  Language Models}.
\newblock In {\em ICML}, 2015.

\bibitem[\protect\citeauthoryear{Jha \bgroup \em et al.\egroup
  }{2010}]{jha&al10}
A.~Jha, V.~Gogate, A.~Meliou, and D.~Suciu.
\newblock {Lifted Inference Seen from the Other Side : The Tractable Features}.
\newblock In {\em NIPS}, 2010.

\bibitem[\protect\citeauthoryear{Kappes \bgroup \em et al.\egroup
  }{2015}]{kappes2013comparative}
J.~Kappes, B.~Andres, A.~Hamprecht, C.~Schn\"{o}rr, S.~Nowozin, D.~Batra,
  S.~Kim, B.~Kausler, T.~Kr\"{o}ger, J.~Lellmann, N.~Komodakis, B.~Savchynskyy,
  and C.~Rother.
\newblock {A Comparative Study of Modern Inference Techniques for Structured
  Discrete Energy Minimization Problems}.
\newblock {\em In IJCV}, 2015.

\bibitem[\protect\citeauthoryear{Kersting \bgroup \em et al.\egroup
  }{2009}]{kersting&al09}
K.~Kersting, B.~Ahmadi, and S.~Natarajan.
\newblock {Counting Belief Propagation}.
\newblock In {\em UAI}, 2009.

\bibitem[\protect\citeauthoryear{Kersting}{2012}]{kersting12}
K.~Kersting.
\newblock {Lifted Probabilistic Inference}.
\newblock In {\em ECAI}, 2012.

\bibitem[\protect\citeauthoryear{Kiddon and Domingos}{2011}]{kiddon&domingos11}
C.~Kiddon and P.~Domingos.
\newblock {Coarse-to-Fine Inference and Learning for First-Order Probabilistic
  Models.}
\newblock In {\em AAAI}, 2011.

\bibitem[\protect\citeauthoryear{Kimmig \bgroup \em et al.\egroup
  }{2015}]{kimmig&al15}
A.~Kimmig, L.~Mihalkova, and L.~Getoor.
\newblock {Lifted Graphical Models: A Survey}.
\newblock {\em Machine Learning}, 2015.

\bibitem[\protect\citeauthoryear{Kohli \bgroup \em et al.\egroup
  }{2013}]{kohli2013principled}
P.~Kohli, A.~Osokin, and S.~Jegelka.
\newblock {A Principled Deep Random Field Model for Image Segmentation}.
\newblock In {\em CVPR}, 2013.

\bibitem[\protect\citeauthoryear{Lempitsky \bgroup \em et al.\egroup
  }{2010}]{5166447}
V.~Lempitsky, C.~Rother, S.~Roth, and A.~Blake.
\newblock {Fusion Moves for Markov Random Field Optimization}.
\newblock {\em In PAMI}, Aug 2010.

\bibitem[\protect\citeauthoryear{Mittal \bgroup \em et al.\egroup
  }{2014}]{mittal&al14}
H.~Mittal, P.~Goyal, V.~Gogate, and P.~Singla.
\newblock {New Rules for Domain Independent Lifted {MAP} Inference}.
\newblock In {\em NIPS}, 2014.

\bibitem[\protect\citeauthoryear{Mittal \bgroup \em et al.\egroup
  }{2015}]{mittal&al15}
H.~Mittal, A.~Mahajan, V.~Gogate, and P.~Singla.
\newblock {Lifted Inference Rules With Constraints}.
\newblock In {\em NIPS}, 2015.

\bibitem[\protect\citeauthoryear{Mladenov \bgroup \em et al.\egroup
  }{2014}]{mladenov&al14}
M.~Mladenov, K.~Kersting, and A.~Globerson.
\newblock {Efficient Lifting of {MAP} LP Relaxations Using k-Locality}.
\newblock In {\em AISTATS}, 2014.

\bibitem[\protect\citeauthoryear{Mozerov and van~de Weijer}{2015}]{TSGO}
M.~G. Mozerov and J.~van~de Weijer.
\newblock {Accurate Stereo Matching by Two-Step Energy Minimization}.
\newblock {\em IEEE Transactions on Image Processing}, March 2015.

\bibitem[\protect\citeauthoryear{Nath and Domingos}{2010}]{nath&domingos10}
A.~Nath and P.~Domingos.
\newblock {Efficient Lifting for Online Probabilistic Inference}.
\newblock In {\em AAAIWS}, 2010.

\bibitem[\protect\citeauthoryear{Nath and Domingos}{2016}]{nath&domingos16}
A.~Nath and P.~Domingos.
\newblock {Learning Tractable Probabilistic Models for Fault Localization}.
\newblock In {\em AAAI}, 2016.

\bibitem[\protect\citeauthoryear{Niepert}{2012}]{niepert12}
M.~Niepert.
\newblock {Markov Chains on Orbits of Permutation Groups}.
\newblock In {\em UAI}, 2012.

\bibitem[\protect\citeauthoryear{Noessner \bgroup \em et al.\egroup
  }{2013}]{noessner&al13}
J.~Noessner, M.~Niepert, and H.~Stuckenschmidt.
\newblock {Rock{I}t: Exploiting Parallelism and Symmetry for {MAP} Inference in
  Statistical Relational Models}.
\newblock In {\em AAAI}, 2013.

\bibitem[\protect\citeauthoryear{Poole}{2003}]{poole03}
D.~Poole.
\newblock {First-Order Probabilistic Inference}.
\newblock In {\em IJCAI}, 2003.

\bibitem[\protect\citeauthoryear{Sarkhel \bgroup \em et al.\egroup
  }{2014}]{sarkhel&al14}
S.~Sarkhel, D.~Venugopal, P.~Singla, and V.~Gogate.
\newblock Lifted {MAP} inference for {M}arkov logic networks.
\newblock In {\em AISTATS}, 2014.

\bibitem[\protect\citeauthoryear{Sarkhel \bgroup \em et al.\egroup
  }{2015}]{sarkhel&al15}
S.~Sarkhel, P.~Singla, and V.~Gogate.
\newblock {Fast Lifted {MAP} Inference via Partitioning}.
\newblock In {\em NIPS}, 2015.

\bibitem[\protect\citeauthoryear{Scharstein and
  Szeliski}{2002}]{Scharstein2002}
D.~Scharstein and R.~Szeliski.
\newblock {A Taxonomy and Evaluation of Dense Two-Frame Stereo Correspondence
  Algorithms}.
\newblock {\em In IJCV}, 2002.

\bibitem[\protect\citeauthoryear{Scharstein and
  Szeliski}{2003}]{Scharstein:2003:HSD:1965841.1965865}
D.~Scharstein and R.~Szeliski.
\newblock {High-accuracy Stereo Depth Maps Using Structured Light}.
\newblock In {\em CVPR}, 2003.

\bibitem[\protect\citeauthoryear{Singla and Domingos}{2008}]{singla&domingos08}
P.~Singla and P.~Domingos.
\newblock {Lifted First-Order Belief Propagation}.
\newblock In {\em AAAI}, 2008.

\bibitem[\protect\citeauthoryear{Singla \bgroup \em et al.\egroup
  }{2014}]{singla&al14}
P.~Singla, A.~Nath, and P.~Domingos.
\newblock {Approximate Lifting Techniques for Belief Propagation}.
\newblock In {\em AAAI}, 2014.

\bibitem[\protect\citeauthoryear{Szeliski \bgroup \em et al.\egroup
  }{2008}]{szeliski&al06}
R.~Szeliski, R.~Zabih, D.~Scharstein, O.~Veksler, V.~Kolmogorov, A.~Agarwala,
  M.~Tappen, and C.~Rother.
\newblock {A Comparative Study of Energy Minimization Methods for Markov Random
  Fields with Smoothness-Based Priors}.
\newblock {\em In PAMI}, June 2008.

\bibitem[\protect\citeauthoryear{Van~den Bergh \bgroup \em et al.\egroup
  }{2012}]{bergh&al12}
M.~Van~den Bergh, X.~Boix, G.~Roig, B.~de~Capitani, and L.~Van~Gool.
\newblock {SEEDS: Superpixels Extracted via Energy-Driven Sampling}.
\newblock In {\em ECCV}, 2012.

\bibitem[\protect\citeauthoryear{Van~den Broeck and
  Darwiche}{2013}]{broeck&darwiche13}
G.~Van~den Broeck and A.~Darwiche.
\newblock {On the Complexity and Approximation of Binary Evidence in Lifted
  Inference}.
\newblock In {\em NIPS}, 2013.

\bibitem[\protect\citeauthoryear{Van~den Broeck and
  Niepert}{2015}]{broeck&niepert15}
G.~Van~den Broeck and M.~Niepert.
\newblock {Lifted Probabilistic Inference for Asymmetric Graphical Models}.
\newblock In {\em AAAI}, 2015.

\bibitem[\protect\citeauthoryear{Venugopal and
  Gogate}{2014}]{venugopal&gogate14evidence}
D.~Venugopal and V.~Gogate.
\newblock {Evidence-Based Clustering for Scalable Inference in Markov Logic}.
\newblock In {\em Joint ECML-KDD}, 2014.

\bibitem[\protect\citeauthoryear{Wei \bgroup \em et al.\egroup
  }{2016}]{WeiXingSH}
X.~Wei, Q.~Yang, Y.~Gong, M.~Yang, and N.~Ahuja.
\newblock {Superpixel Hierarchy}.
\newblock {\em CoRR}, abs/1605.06325, 2016.

\end{thebibliography}
}
\end{document}